\documentclass[journal,twoside,web]{ieeecolor} 
\usepackage{tmi}
\usepackage{cite}
\usepackage{amsmath,amssymb,amsfonts}
\usepackage{textcomp}

\usepackage{times}
\usepackage{epsfig}
\usepackage{graphicx}
\let\labelindent\relax
\usepackage{enumitem}
\usepackage{colortbl}
\usepackage{epstopdf}
\usepackage{subfigure}
\usepackage{algorithm}
\usepackage{algpseudocode}
\usepackage{boldline}
\usepackage{balance}
\usepackage{booktabs}
\definecolor{row_color}{rgb}{.92,.96,.96}
\newcommand{\bfsection}[1]{\vspace*{0.1cm}\noindent\textbf{#1:}}

\newcommand{\red}[1]{\textcolor{red}{#1}}

\newcommand{\etal}{\textit{et al.}}
\newcommand{\eg}{\textit{e.g.}}

\def\BibTeX{{\rm B\kern-.05em{\sc i\kern-.025em b}\kern-.08em
    T\kern-.1667em\lower.7ex\hbox{E}\kern-.125emX}}
\begin{document}
\title{Learning Vector Quantized Shape Code for Amodal Blastomere Instance Segmentation}

\author{Won-Dong Jang$^{*}$, \and Donglai Wei, \and Xingxuan Zhang, \and Brian Leahy, \and Helen Yang, \and James Tompkin, \and Dalit Ben-Yosef, \and Daniel Needleman, \and and Hanspeter Pfister 
\thanks{W. Jang, D. Wei, B. Leahy, H. Yang, H. Pfister, D. Needleman are with the School of Engineering and Applied Sciences, Harvard University, Cambridge, MA, USA.
corresponding author email: wdjang@g.harvard.edu}
\thanks{X. Zhang is with the School of Engineering, Jiaotong University, China.}
\thanks{J. Tompkin is with the Department of Computer Science, Brown University, Cambridge, MA, USA.}
\thanks{D. Ben-Yosef is with the Tel Aviv Sourasky Medical Center}
}

\maketitle
\begin{abstract}
Blastomere instance segmentation is important for analyzing embryos' abnormality. To measure the accurate shapes and sizes of blastomeres, their amodal segmentation is necessary.
Amodal instance segmentation aims to recover the complete silhouette of an object even when the object is not fully visible. 
For each detected object, previous methods directly regress the target mask from input features.
However, images of an object under different amounts of occlusion should have the same amodal mask output, which makes it harder to train the regression model. To alleviate the problem, we propose to classify input features into intermediate shape codes and recover complete object shapes from them. 
First, we pre-train the Vector Quantized Variational Autoencoder (VQ-VAE) model to learn these discrete shape codes from ground truth amodal masks. Then, we incorporate the VQ-VAE model into the amodal instance segmentation pipeline with an additional refinement module. We also detect an occlusion map to integrate occlusion information with a backbone feature. As such, our network faithfully detects bounding boxes of amodal objects.
On an internal embryo cell image benchmark, the proposed method outperforms previous state-of-the-art methods.
To show generalizability, we show segmentation results on the public KINS natural image benchmark. 
To examine the learned shape codes and model design choices, we perform ablation studies on a synthetic dataset of simple overlaid shapes. Our method would enable accurate measurement of blastomeres in in vitro fertilization (IVF) clinics, which potentially can increase IVF success rate.
\end{abstract}


\section{Introduction}

\IEEEPARstart{I}{nfertile} couples worldwide use In-Vitro Fertilization (IVF) to treat their infertility. In a typical IVF treatment, clinicians stimulate the woman to produce many eggs, fertilize those eggs, and culture the resulting embryos for 3--5 days. The clinicians then visually inspect the embryos, select the one that appears most likely to form a viable pregnancy, and transfer it back to the mother. To aid in embryo selection, many modern clinics employ sophisticated time-lapse imaging systems~\cite{armstrong2019Timelapse} that record three-dimensional videos of the embryos as they develop. 

One feature known to be predictive of an embryo's viability is the shape and symmetry among the cells in the early developing embryo, which are known as blastomeres~\cite{racowsky2011National}. However, current clinical practice is to visually score the symmetry at a few distinct points in time, which is time-consuming, inaccurate, and omits much information about the embryo, especially when time-lapse imaging is used. This makes replacing visual symmetry scoring with automated blastomere segmentation a prime candidate for improving clinical IVF practice.

However, while clinics have collected a lot of embryo images from IVF cycles, most existing blastomere segmentation algorithms~\cite{rad2018hybrid,sidhu2019Automated,moradirad2019CellNet,kheradmand2019Preimplantation,khan2016Segmentation} use hand-crafted features instead of data-driven approaches. Since hand-crafted methods are tailored to a certain dataset, they may not be robust on different datasets that are collected in varying environments. In this work, we propose a convolutional neural network, which performs amodal visual reconstruction for blastomere segmentation.

\begin{figure}[t]
    \centering
    \includegraphics[width=\columnwidth]{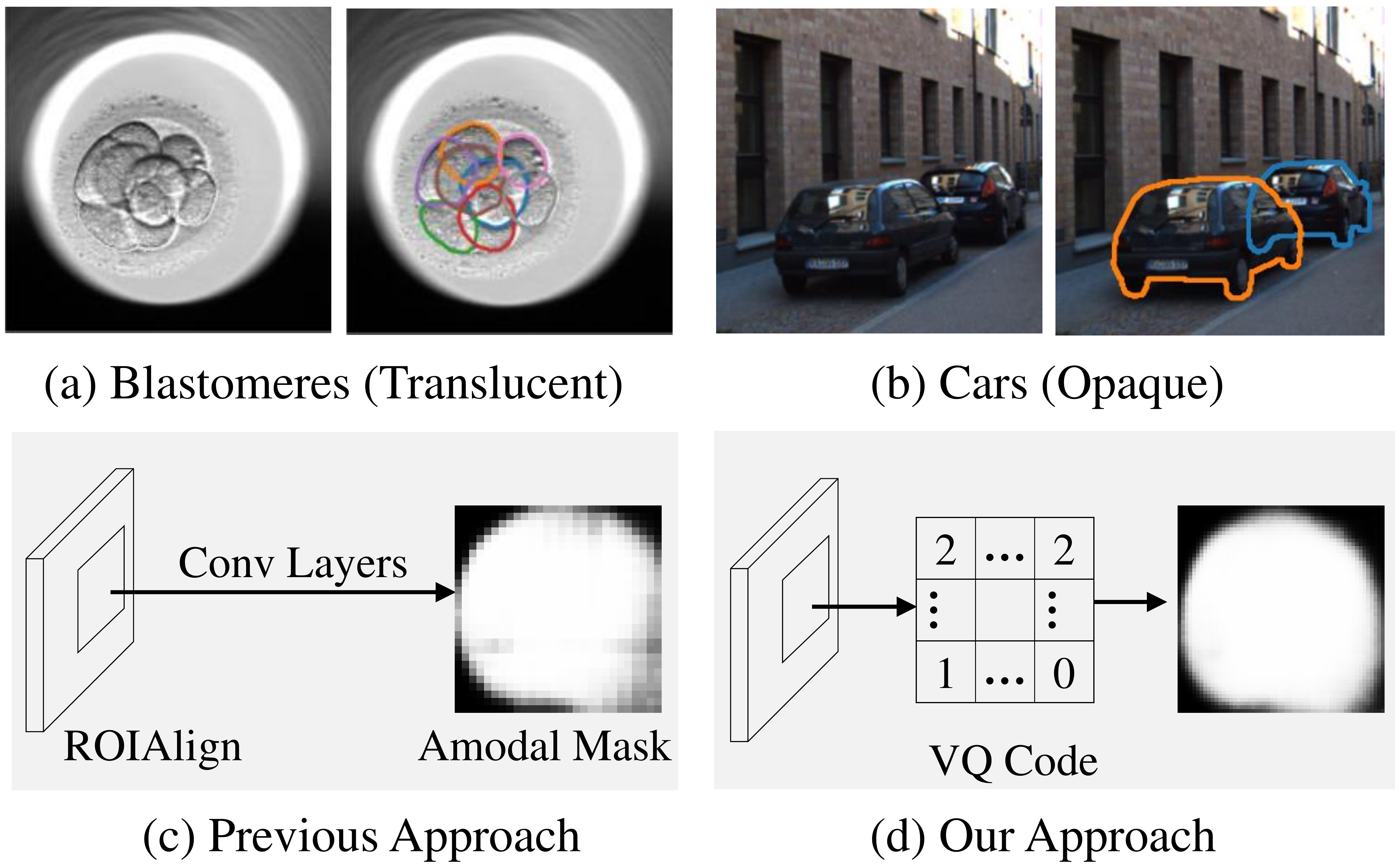}
    \caption{\textbf{Amodal instance segmentation.} 
    We show an image and its amodal segmentation mask for two common cases:
    (a) translucent objects overlapping with each other, and 
    (b) opaque objects occluding each other.
    (c) Previous approaches directly regress the amodal mask from the region of interest (ROIAlign) features. (d) Instead, we first learn a vector quantized (VQ) shape code from ground truth amodal masks, and then classify ROIAlign features into these discrete codes.}
    \label{fig:motivation}
\end{figure}


Amodal visual reconstruction, predicting the complete shape of partially-visible objects, is part of human ordinary perception. Two common examples of this are: 1) translucent objects visually overlap within the camera's view, such as when observing biomedical images of cells (Fig.~\ref{fig:motivation}\red{a}), and 2) opaque objects occlude each other and only a portion of the object is visible, such as when looking down a street at a row of parked cars (Fig.~\ref{fig:motivation}\red{b}). Beyond its importance in cognitive psychology, amodal visual perception can greatly benefit computer vision applications in practice. With it, biologists can examine new hypothesis through automatic large-scale cell shape measurement from light microscopy images and robotic agents can better navigate through complex environments with partially visible objects.

Unlike the typical instance segmentation setting, which only requires us to label the visible pixels~\cite{li2017fully,he2017mask,liu2018path,dai2016instance}, our wish to predict the shape for invisible or partially-visible object regions requires us to fit a model of shape to the image. Classic solutions have tried known rigid templates of the target object~\cite{knoll1986recognizing}, statistical models which capture object shape variation~\cite{Cootes1995}, or discriminative parts-based models learned from a dataset~\cite{parizi2014automatic} potentially with explicit occlusion reasoning~\cite{winn2006layout}. 

Many recent deep-learning-based models have been proposed for amodal segmentation~\cite{li2016amodal,zhu2017semantic,ehsani2018segan,follmann2019learning,hu2019sail,qi2019amodal}. However, these approaches often do not have prior knowledge of the underlying shape, which makes the shape difficult to predict from instance observations under different amounts of occlusion. Further, unlike normal instance segmentation, images of an object under different amounts of occlusion should have the same amodal mask output. Thus, it will be more robust to classify input features into an intermediate robust representation instead of working on the pixel-level.

To exploit this additional information, we propose to learn discrete supervised learning amodal instance segmentation algorithm for partially-visible objects. From binary masks of our object class, we create a deep shape prior as an embedding space with a vector quantized-variational autoencoder (VQ-VAE;~\cite{kingma2013auto}). Then, we train our segmentation model to predict the latent representation of an object mask in a bounding box.

Segmentation performance of proposal-based instance segmentation methods~\cite{he2017mask,liu2018path} highly depends on the bounding box quality. In amodal segmentation, occlusion makes having accurate bounding boxes even more difficult. To tackle this occlusion problem, we add an occlusion detection module to a backbone network. This allows our network to propose better bounding boxes by integrating the occlusion information with the backbone features.

We experiment with a real embryo cell biomedical dataset. Furthermore, we conduct experiments on a synthetic dataset and natural images of street scenes via the KINS dataset~\cite{alhaija2018augmented} to show generalizability of our method. Our approach of encoding objects outperforms state-of-the-art instance segmentation algorithms~\cite{he2017mask,qi2019amodal} on both the translucent and occluded types of tested partial visibility.

In summary, our contribution is to propose a novel formulation that incorporates a vector quantized shape code into the amodal instance segmentation pipeline. Additionally, we exploit occlusion information when detecting and segmenting amodal objects via occlusion detection, which can be a new direction for amodal segmentation. This method achieves state-of-the-art performance on not only an internal biomedical image dataset but also the KINS natural image dataset. 
Finally, to the best of our knowledge, this is the first approach that applies amodal instance segmentation method to blastomere segmentation.


\section{Related Works}
\bfsection{Blastomere Segmentation}
Traditional methods predict semantic blastomere masks using hand-crafted features without the instance-level segmentation.
Khan~\etal~\cite{khan2016Segmentation} set seeds inside and outside of cells and optimize Markov random field for segmentation. 
Rad~\etal~\cite{rad2018hybrid} and Kheradmand~\etal~\cite{kheradmand2019Preimplantation} generate blastomere candidates from extracted edges and select the best candidate with in terms of edge coverage.
Sidhu and Mills~\cite{sidhu2019Automated} apply thresholding and morphological operations to find the regions of blastomeres and find centers of each cell by measuring distances from pixels to the closest boundary.

Cell-Net proposed by Rad~\etal~\cite{moradirad2019CellNet} is the closest method to ours, training a convolutional neural network for cell localization. However, Cell-Net only predicts blastomere centers, while we perform amodal instance segmentation.

\bfsection{Amodal Instance Segmentation}
Partially-visible object segmentation is typically studied in biomedical image analysis where cells are often translucent. For nuclei segmentation, Molnar~\etal~\cite{molnar2016accurate} fit a circular active contour model~\cite{Kass1988} using multiple layered distributions of the number of nuclei per pixel. Plissiti and Nikou~\cite{plissiti2012overlapping} segment overlapping nuclei by combining nuclei boundary features with priori knowledge of nuclei shape. However, the proposed method works only when given two nuclei centers and requires parameter tuning. Lee and Kim~\cite{LeeKim2016} approach translucent cell data as a problem of superpixel segmentation for seed location, and of contour attribution and refinement via graph cuts. B\"{o}hm et al.~\cite{Bohm2018} segment translucent cell data by learning to lift the image into 3D via a UNet architecture. In both cases, the shape of the object is not specifically represented (e.g., implicitly via a prior), which makes handling occlusion-based partial visibility difficult.

Some works exist on more natural images, e.g., Kihara~\etal~\cite{kihara2016shadows} exploit occlusion as a signal to recover full masks for object instances via a Shape Boltzmann machine~\cite{eslami2014shape}, but not for translucent objects. 
Li and Malik~\cite{li2016amodal} introduce the first amodal segmentation method. They predict bounding-boxes of modal parts of objects using the object detector~\cite{ren2015faster} and extract segmentation masks using a neural network accepting a pair of an image and a bounding-box as the input. The proposed algorithm iteratively updates segmentation masks by recomputing the bounding-boxes from the output of the network. Zhu~\etal~\cite{zhu2017semantic} announce datasets for class-independent amodal segmentation. Multiple subjects annotate the BSDS dataset to analyze the consistency between them. For computational model comparison, they evaluate modal and amodal object proposal algorithms on the proposed amodal COCO dataset. Ehsani~\etal~\cite{ehsani2018segan} first perform amodal segmentation and then apply a generative adversarial network to have a complete object image by synthesizing the amodal area. Follman~\etal~\cite{follmann2019learning} predict amodal masks as well as visible masks for occlusion reasoning. Hu~\etal~\cite{hu2019sail} present a synthetic dataset for amodal instance level video object segmentation. Qi~\etal~\cite{qi2019amodal} present an amodal segmentation dataset, KINS, by annotating the KITTI detection dataset. They also propose an amodal segmentation network by adding occlusion classification and amodal segmentation branches to the Mask R-CNN framework~\cite{he2017mask}.

Recently, Isack~\etal~\cite{isack2018k} introduce the notion of K-convexity, and demonstrated its application in translucent instance segmentation via an energy minimization on an MRF. This allows enforcing a convexity prior on the shape of an instance (such as star~\cite{veksler2008star}, geodesic-star~\cite{gulshan2010geodesic}, hedgehog~\cite{isack2016hedgehog}, or regular~\cite{gorelick2017convexity}). However, K-convexity optimization requires seed annotation for each object instance. In contrast, our method learns pixel-wise shape priors and does not require seed annotations. 

\begin{figure}[t]
    \centering
    \includegraphics[width=\columnwidth]{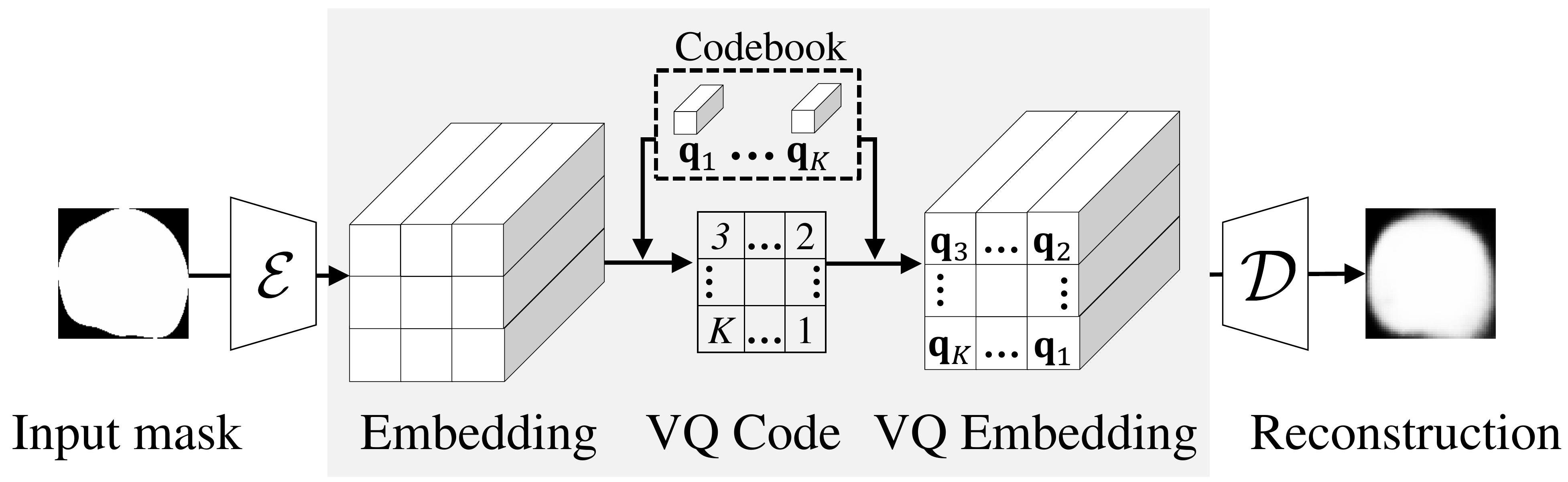}
    \caption{VQ-VAE architecture containing the mask encoder, embedding quantizer, and mask decoder networks.}
    \label{fig:vqvae}
\end{figure}

\bfsection{Deep Learning for Shape Prior}
These approaches are common in 3D shape completion. Wu~\etal~\cite{wu2016single} reconstruct 3D shapes by training shape priors from 3D skeleton parameters; they also later consider the naturalness of reconstructed shapes when training shape priors~\cite{wu2018learning}. Dai~\etal~\cite{dai2017shape} transform incomplete 3D scans into complete 3D shapes by learning from template shapes. Stutz and Geiger~\cite{stutz2018learning} adopt a variational autoencoder~\cite{kingma2013auto}. For detection-based instance object segmentation, Kuo~\etal~\cite{kuo_shapemask_2019} construct a set of prior masks for each object class and align one of the templates within a bounding box to use it as a shape prior for mask generation. However, deep shape priors are less common for amodal segmentation, where objects overlap each other. 

\bfsection{Deep Learning for Vector Quantization}
Vector quantization methods have been widely used for image compression~\cite{agustsson2017soft,theis2017lossy}. Recently, van den Oord \etal~\cite{van2017neural} proposed a vector quantized variational autoencoder for image generation. They show that the proposed method generates more realistic images using learned template codewords. Based on the vector quantized variational autoencoder, Razavi \etal~\cite{razavi2019generating} developed a hierarchical autoencoder, which encodes an input image in high and low levels. While the high-level codewords contain global information, the low ones have local features. The hierarchical method synthesizes high-quality images by utilizing both global and local information.

\section{Vector Quantized Shape Code}
\label{sec:vq_vae}

Our goal is to learn a discrete representation of amodal shape masks. With it, we can re-formulate the amodal instance segmentation as a classification problem in the low-dimensional latent space. Comparing to previous dense pixel-level mask prediction, the proposed approach can be robust to occlusion changes and regularized in geometry.
To this end, we train a vector quantized variational autoencoder (VQ-VAE) model on the amodal masks to learn the vector quantized (VQ) shape code.

\bfsection{Comparing Latent Variable Models}
To learn a compact representation of the input,
variational autoencoder models (VAE)~\cite{kingma2013auto} are commonly used with the Gaussian prior distribution of the latent variable. 
VAEs learn a global continuous code of the input with the mask encoder model $\mathcal{E}$, which can be decoded back for input reconstruction with the mask decoder model $\mathcal{D}$.
To discretize the learned code, VAE-based clustering methods jointly learn a codebook of embedding vectors that serve as clustering centers. However, as the learned embedding is global, it takes a large codebook for the input to find a similar quantized code. It requires an even larger codebook for a larger number of object categories.
VQ-VAEs~\cite{van2017neural} predict embeddings with spatial resolution and jointly learn a global codebook (Fig.~\ref{fig:vqvae}). With it, we can use the quantized embeddings to reconstruct input with a limited codebook size.

\bfsection{VQ-VAE Model}
The key component of VQ-VAE models is the embedding quantizer module. During inference, the mask encoder first transforms the input binary mask $\mathbf{x}$ into a set of latent vectors $\mathbf{e}$. Then, the embedding quantizer assigns each latent vector to the nearest code in the pre-trained codebook $\{\mathbf{q}_1,\ldots, \mathbf{q}_K\}$. Lastly, the mask decoder transforms the quantized embeddings $\mathbf{\hat{e}}$ back into a binary mask. 

\bfsection{Learning}
The loss function combines a reconstruction loss, a codebook loss, and a commitment loss. The reconstruction loss is defined as the cross-entropy loss between input mask $\mathbf{x}$ and the reconstructed mask $\mathcal{D}(\mathbf{\hat{e}})$.
The codebook loss, which only applies to the codebook, makes the selected codes $\mathbf{\hat{e}}$ close to the predicted latent vector $\mathbf{e}$. The commitment loss, which only applies to the mask encoder, forces the latent vectors $\mathcal{E}(\mathbf{x})$ to stay close to the matched codes to prevent excessive fluctuations of codes. The full VQ-VAE loss function $\mathcal{L}_\textrm{v}$ is
\begin{align}
    \mathcal{L}_\textrm{v}
    =\|\mathbf{x}-\mathcal{D}(\mathbf{\hat{e}})\|_{2}^{2}
    +\|[\mathbf{e}]-\mathbf{\hat{e}}\|_{2}^{2}
    +\beta&\|\mathcal{E}(\mathbf{x})-[\mathbf{\hat{e}}]\|_{2}^{2},
\end{align}
where the operator $[.]$ stands for a stop gradient operation that blocks gradients from flowing into its argument, and $\beta$ is a hyper-parameter, which is set to 0.25.



\bfsection{Implementation Details}
The mask encoder has three convolution layers, two residual modules, and one convolution layer. The stride for each convolution layer is 2, which reduces the spatial resolution by half at each layer. 
For the three convolutional layers, we use 32, 64, and 128 $4 \times 4$ sized filters, respectively.
Thus, the mask encoder changes the spatial resolution from $H \times W$ to $H/8 \times W/8$. In the last convolution layer, we set the embedding dimension to 16 empirically. Hence, the mask encoder yields a $H/8 \times W/8 \times 16$ tensor, which is a set of 16-dimensional latent vectors in embedding space. 

For the embedding quantizer, we set the number of codewords $K$ to 4 empirically, as binary masks are much easier to model than natural images. Also, $K$ codewords have $K \times H/8 \times W/8$ possible combinations, which is large enough to model binary object masks.

The mask decoder has one convolutional layer, two residual modules, and three deconvolutional layers. 
Note that each axis of the input image is reconstructed to its original size via the deconvolutional layers. At the end of the decoder, we add a sigmoid layer to constrain values in the reconstructed masks ranging from 0 to 1. The VQ-VAE model is trained separately, and its parameters are fixed after training. 
\section{Amodal Instance Segmentation Pipeline}\label{sec:method}
\begin{figure}[t]
    \centering
    \includegraphics[width=\columnwidth]{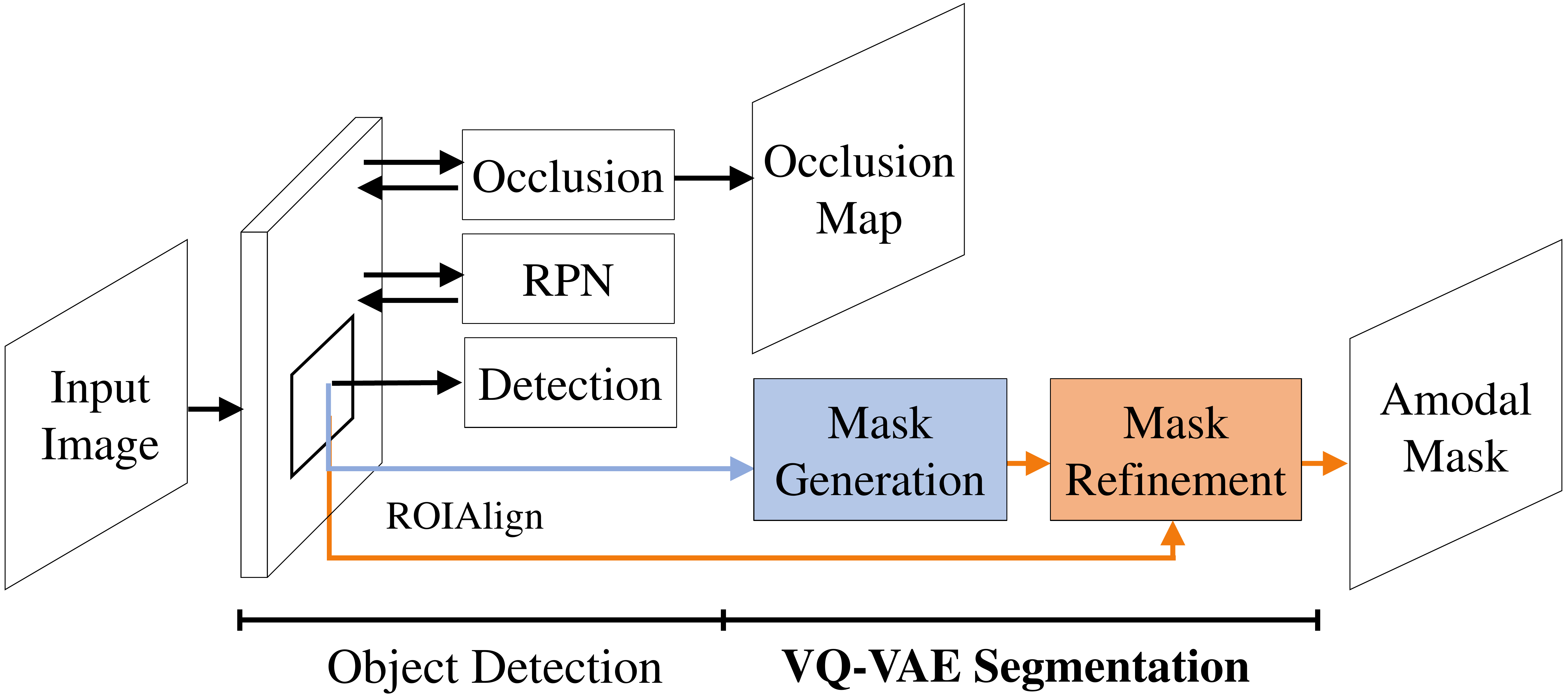}
    \caption{\textbf{Overview of amodal segmentation pipeline.} 
    We start from an instance segmentation pipeline, \eg, Mask-RCNN. We add the occlusion detection module and replace the original FCN with the proposed VQ-VAE segmentation module. The proposed segmentation model has two steps: initial mask generation through VQ shape code prediction and mask refinement for better localization.
    }
    \label{fig:overview}
\end{figure}

We propose the VQ-VAE segmentation module to improve amodal instance segmentation. We take the proposal-based instance segmentation approach that contains two modules: object detection and mask prediction (Fig.~\ref{fig:overview}). 
We attach an occlusion detection branch to object detection (Sec.~\ref{subsec:objectdetection}) and replace previous fully convolutinal network (FCN) with the proposed module for mask prediction (Sec.~\ref{subsec:method_seg}).
The whole pipeline is trained end-to-end (Sec.~\ref{subsec:method_learn}).

\subsection{Object Detection Module}
\label{subsec:objectdetection}
 
\bfsection{Backbone} To extract effective features from the input image, we use ResNet50~\cite{he2016deep} as a backbone network, which is trained using the ImageNet dataset~\cite{deng2009imagenet}. We drop the average pooling layer from the ResNet50 to use spatial features. 

\bfsection{Bounding Box Detection}
The region proposal network (RPN)~\cite{ren2015faster} takes features from the backbone network and measures object existence probabilities and regression parameters of bounding boxes. We employ feature pyramidal networks to extract features across five scales, and minimize the sum of the loss functions at all scales. For each region of interests (ROI), we predict regression parameters and classify its object category. 

\bfsection{Occlusion Detection}
Unlike Faster-RCNN~\cite{ren2015faster}, our detection module predicts both bounding boxes and a binary occlusion map. Detecting locations of occlusions allows our object detection module to predict accurate bounding boxes for partially visible objects. Using the backbone features, we estimate probabilities of each pixel being occluded $\{d_{i}\}$ via four convolution layers. We adopt the binary cross entropy loss:
\begin{equation}
    \mathcal{L}_\textrm{o}=-\sum_{i \in H \times W}{\left\{l_{i} \log d_{i}+\left(1-l_{i}\right) \log \left(1-d_{i}\right)\right\}},
\end{equation}
where $H \times W$ is the spatial resolution of the backbone feature map and $l_{i}$ is the ground-truth occlusion label at pixel $i$. We concatenate the output of the second-to-last convolution layer and the backbone feature map to exploit occlusion information in the detection and segmentation modules.

\begin{figure}[t]
    \centering
    \includegraphics[width=\columnwidth]{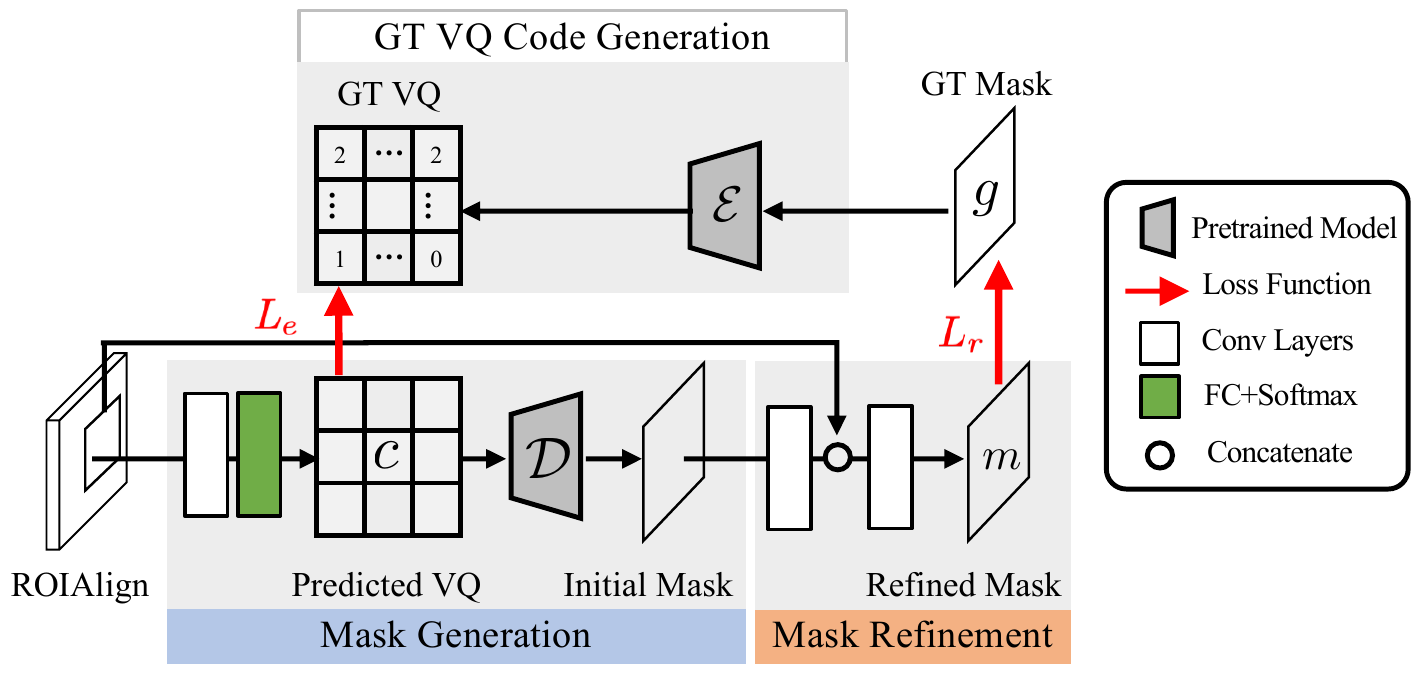}
    \caption{\textbf{VQ-VAE segmentation module.} We have two segmentation stages: mask generation and mask refinement. We simultaneously minimize the two loss functions, $\mathcal{L}_\textrm{e}$ and $\mathcal{L}_\textrm{r}$.}
    \vspace{-0.25cm}
    \label{fig:maskGeneration}
\end{figure}



\subsection{VQ-VAE Segmentation Module}\label{subsec:method_seg}
As shown in Fig~\ref{fig:maskGeneration}, the proposed VQ-VAE segmentation module has two steps: initial mask generation and mask refinement. It first generates an initial mask through decoding the predicted VQ-VAE shape code. Then, the refinement step learns to better align the initial mask with the visible object boundaries.

\bfsection{Initial Mask Generation}
Given the instance-level feature from the object detection module, we first predict the vector quantized shape code and use a pre-trained VQ-VAE decoder model to decode it into object masks with complete shapes. 

We first predict a vector quantized shape code instead of a pixel-level binary mask to capture complete shapes using VQ-VAE. We use three convolution layers and one fully connected layer to predict codewords of vector quantized shape code $\mathbf{c}$. We formulate the problem of vector quantized shape code prediction as a classification problem. For the classification target, we use the pre-trained VQ-VAE mask encoder $\mathcal{E}$ to encode the ground truth instance mask $\mathbf{g}$ as shown in the right block in Fig~\ref{fig:maskGeneration}. One hot encoding makes the encoded mask $\mathcal{E}(\mathbf{g})$ as a binary representation $\mathbf{b}$. For the codeword classification at each spatial location, the binary cross entropy loss is defined as
\begin{equation}
    \mathcal{L}_\textrm{e}=-\sum_{i \in M \times M \times K }{\left\{b_{i} \log c_{i}+\left(1-b_{i}\right) \log \left(1-c_{i}\right)\right\}},
\end{equation}
where $M \times M$ is a spatial resolution of a vector quantized shape code and $K$ is the number of codewords.
 
We then feed the predicted VQ shape code $\mathbf{c}$ into the VQ-VAE mask decoder $\mathcal{D}$ to obtain an initial mask.

\bfsection{Mask Refinement}\label{subsec:method_refine}
The vector quantized shape code can be powerful for shape completion, but the initial mask may not be well-aligned with the detailed object boundary.
We add another mask refinement step that combine the instance-level feature and the initial mask feature.
To train the refinement decoder, we set its loss function as
\begin{equation}
    \mathcal{L}_\textrm{r} = -\sum_{i\in N \times N \times C}{w_i\left\{g_i \log(m_i) + (1 - g_i) \log(1 - m_i)\right\}},
\end{equation}
where $m_i$ is the probability of a target object occurring at pixel $i$. $N \times N$ is a spatial resolution of the output mask and $C$ indicates the number of object categories. The weight $w_i$ is 1 for the channel of the ground-truth object class, otherwise 0. 

\subsection{Learning Strategy}\label{subsec:method_learn}
During training, parameters in the region proposal network, detection, mask generation, and refinement modules are updated together to minimize the sum of the loss functions: $\mathcal{L}=\mathcal{L}_\textrm{p}+\alpha\mathcal{L}_\textrm{d}+\beta\mathcal{L}_\textrm{o}+\gamma\mathcal{L}_\textrm{e}+\delta\mathcal{L}_\textrm{r}$, where $\mathcal{L}_\textrm{p}$ and $\mathcal{L}_\textrm{d}$ indicate the losses for the region proposal network and the detection module, respectively. Hence, we train the proposed network in an end-to-end manner. Empirically, we set the hyper-parameters $\alpha=\gamma=\delta=1$ and $\beta=0.01$.
\subsection{Implementation Details}

We provide implementation details of the proposed algorithm including architectures and learning strategies.

\bfsection{Architecture}
We shrink the spatial resolution of the initial mask using two convolution layers with strides 2. The refinement network consists of four convolution layers and two deconvolution layers. It outputs class-wise masks at each output channel to decouple segmentation and classification.

\bfsection{Learning}
We initialize parameters in the proposed networks with random values except for the backbone network, which uses weights from the ResNet50~\cite{he2016deep}. We train the network via the stochastic gradient descent optimizer. We set the initial learning rate to $0.04$, and reduce it to $0.004$ and $0.0004$ after 10,000 and 11,000 iterations, respectively. We train networks for 12,000 iterations. We use a minibatch size of 16. It takes less than two days to train the proposed networks.

\bfsection{Running time}
We measure the average computational time of the proposed algorithm on a single Titan X GPU. For images whose shorter axis is fixed to 800, the average running time is 1.75 frames per second. Note that we set the number of proposals to 1,000.

\section{Experiments}
We compare the proposed method with state-of-the-art methods on a microscopy image dataset and a natural image dataset. Then, we perform ablation studies on the natural dataset to better understand each component and to validate our design choices.

\subsection{Experiment Setup}

\bfsection{Comparison methods}
For amodal instance segmentation, we can use different object detection pipelines, \eg, Mask-RCNN~\cite{he2017mask}.
With the same pipeline, the proposed VQ-VAE segmentation module is compared with the fully convolutional network (FCN) on two datasets. 

\bfsection{Metrics} We use mean average precision (mAP), which is standard for object instance segmentation~\cite{lin2014microsoft}. Let $\textrm{AP}_{k}$ denotes a predicted segmentation as correct if its mask intersection over union (IoU) is higher than $k$. 
mAP score is the average of $\{\textrm{AP}_{k}\}$ where $k$ ranges from 0.5 to 0.95 at 0.05 intervals.

\begin{figure}[t]
    \centering
    \includegraphics[width=\columnwidth]{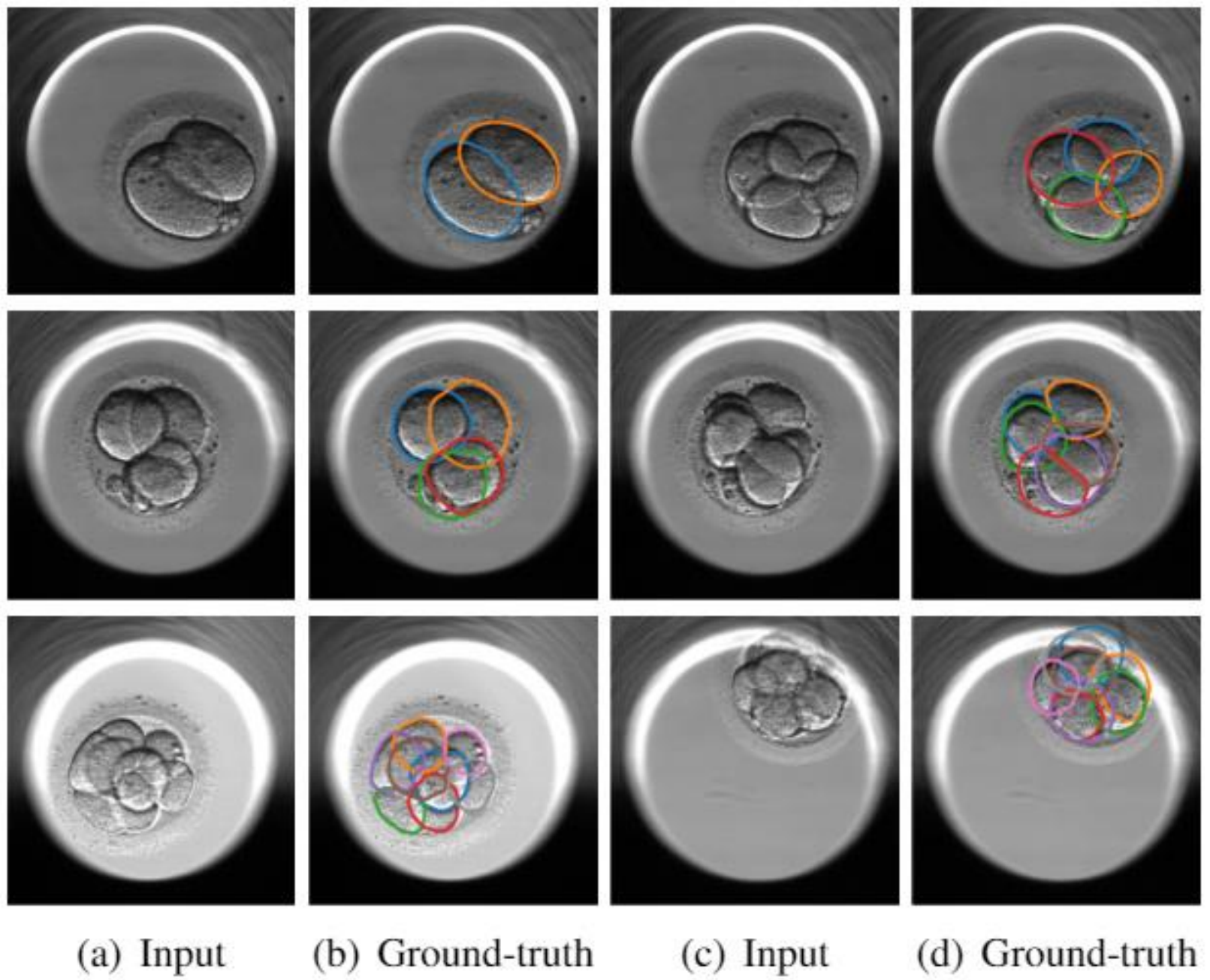}
    \caption{\textbf{Examples of cells in the embryo dataset.} Cells are delin-eated by boundaries with different colors.}
    \label{fig:embryo_data}
\end{figure}

\subsection{Main Results on Embryo Cell Images}


In vitro fertilization clinicians predict embryo transfer success by visually observing cell properties like size, granularity, and cleavage (cell split) timing. Cell segmentation of embryo images would automate this property collection for more efficient prediction. Note that our method is more interpretable by clinicians compared to predicting a single number (cell count) from the input image~\cite{khan_deep_2016}.

\bfsection{Data}
From the IVF clinic in Tel Aviv Medical Center, Israel, we collect 11,671 embryo images, each with a spatial resolution of 500$\times$500 pixels. The numbers of cells in each embryo image varies from 2 to 8. Note that we exclude one cell images to evaluate amodal instance segmentation methods. To obtain ground-truth segments, we annotate cells and then ask experts to proofread the annotations. We use 7,054 images for training and the remaining 4,617 for testing. 
Fig.~\ref{fig:embryo_data} show examples of embryos and their ground-truth annotations for blastomere instances. 
We observe that cells are highly overlapping and only partially visible. The size of cells varies as cells cleave and shrink.


\bgroup
\def\arraystretch{1}
\begin{table}[t]
    \centering
    \caption{Comparison of mAP metric on the embryo cell dataset.}
    \vspace{0.5em}
    \begin{tabular}{lcc}
    \hline 
    Detection& ~FCN~\cite{he2017mask}~&~VQ-VAE (ours)~\\
    \hline
    Mask R-CNN&  0.649 & \bf{0.665} \\
    \hline
    \end{tabular}
\label{table:embryo}
\end{table}
\egroup
\begin{figure*}[t]
    \centering
    \includegraphics[width=\textwidth]{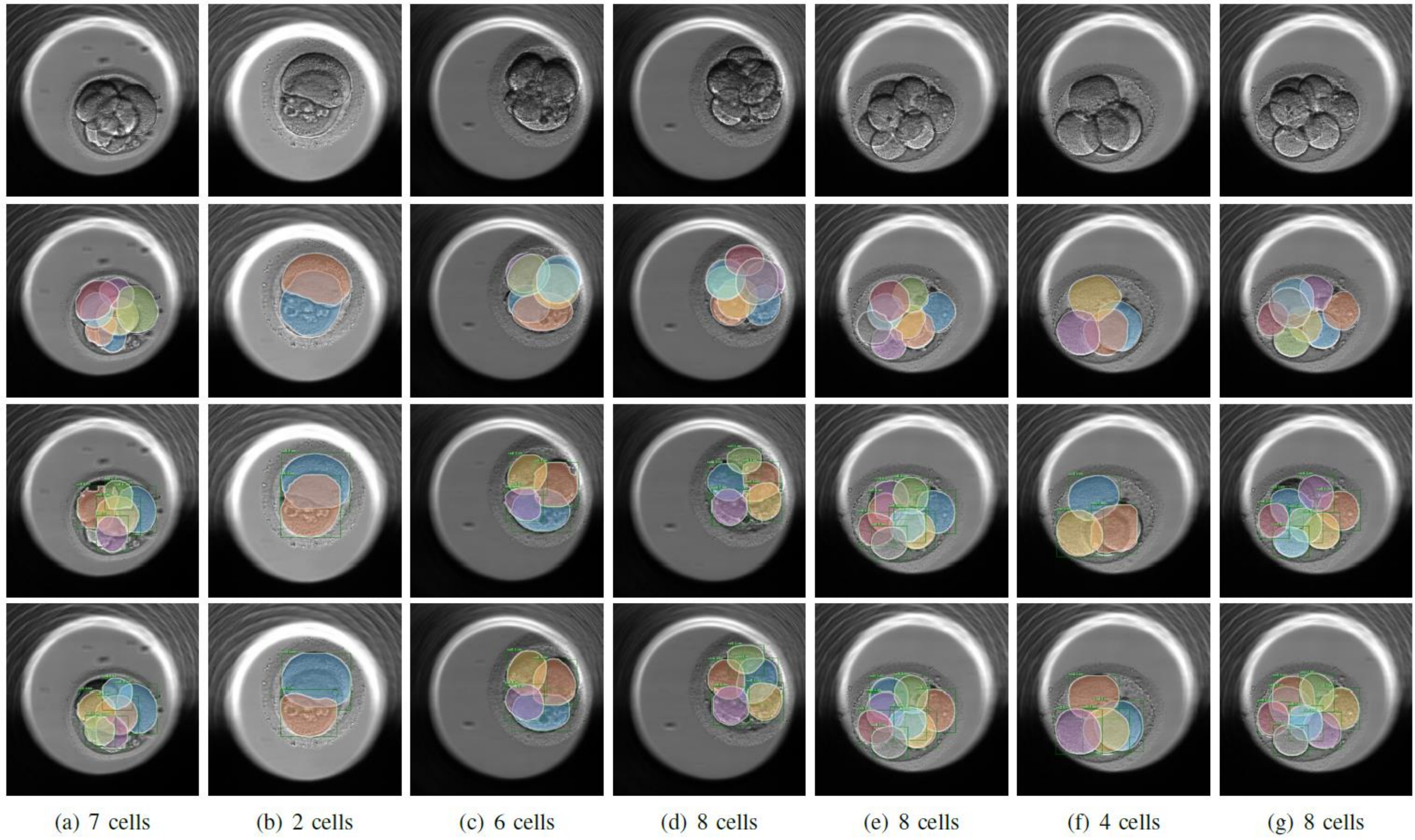}
    \caption{\textbf{Results on embryo cell dataset.} From top to bottom rows, we visualize input images, ground-truth cell masks, and results of FCN and the proposed VQ-VAE, respectively. The segmented object masks are highlighted in coloured regions.}
    \label{fig:embryo_result}
\end{figure*}

\bfsection{Results}
Table~\ref{table:embryo} compares the results of our proposed algorithm with Mask R-CNN~\cite{he2017mask}. We report mean average precision metrics for the evaluation of the cell segmentation methods. The proposed algorithm outperforms the baseline methods. Qualitatively, we observe that the proposed network faithfully detect embryo cells (Fig.~\ref{fig:embryo_result}). Even though partial boundaries of cells are missing, the proposed algorithm generates masks accurately by considering the shape prior of embryo cells.

\bgroup
\def\arraystretch{1}
\begin{table}[t]
    \vspace{0.5cm}
    \centering
    \caption{Comparison of mAP indices on the synthesized dataset.} 
    \vspace{0.1cm}
    \begin{tabular}{lcccc}
    \hline
    Detection& ~FCN~\cite{he2017mask}~ & ~VAE~ & ~VQ-VAE (ours)~\\
    \hline
    Mask R-CNN & 0.809 & 0.849 & \textbf{0.865}\\
    \hline
    \end{tabular}
\label{table:synthesized}
\end{table}
\egroup

\subsection{Additional Results on Synthetic Images}
\begin{figure}[t]
    \centering
    \includegraphics[width=\columnwidth]{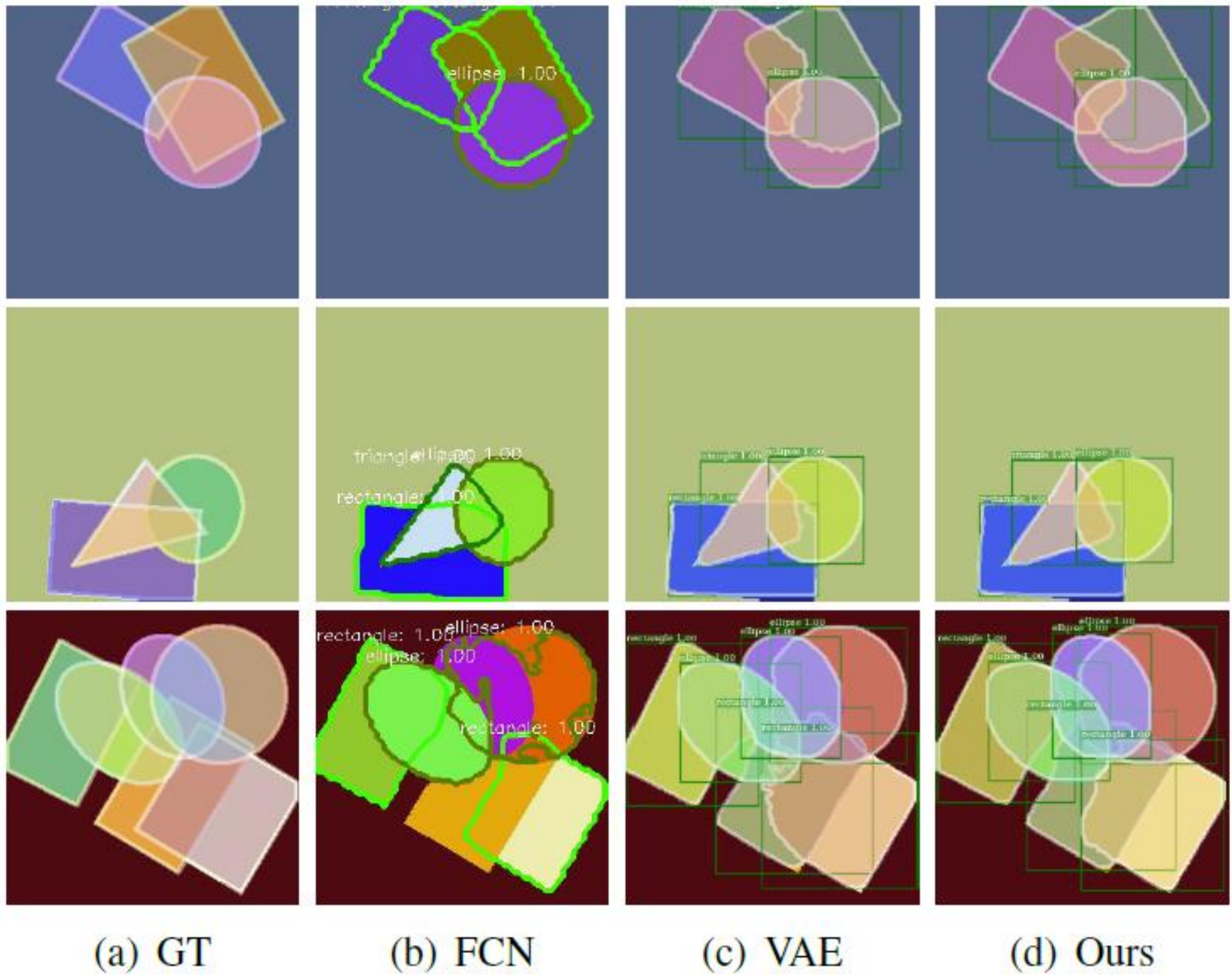}
    \caption{Qualitative comparison of the proposed algorithm to other mask generation methods on the synthesized dataset. Instances are shown with different colors.}
    \label{fig:synthesized}
\end{figure}

\begin{figure}[t]
    \centering
    \scriptsize
    \includegraphics[width=\columnwidth]{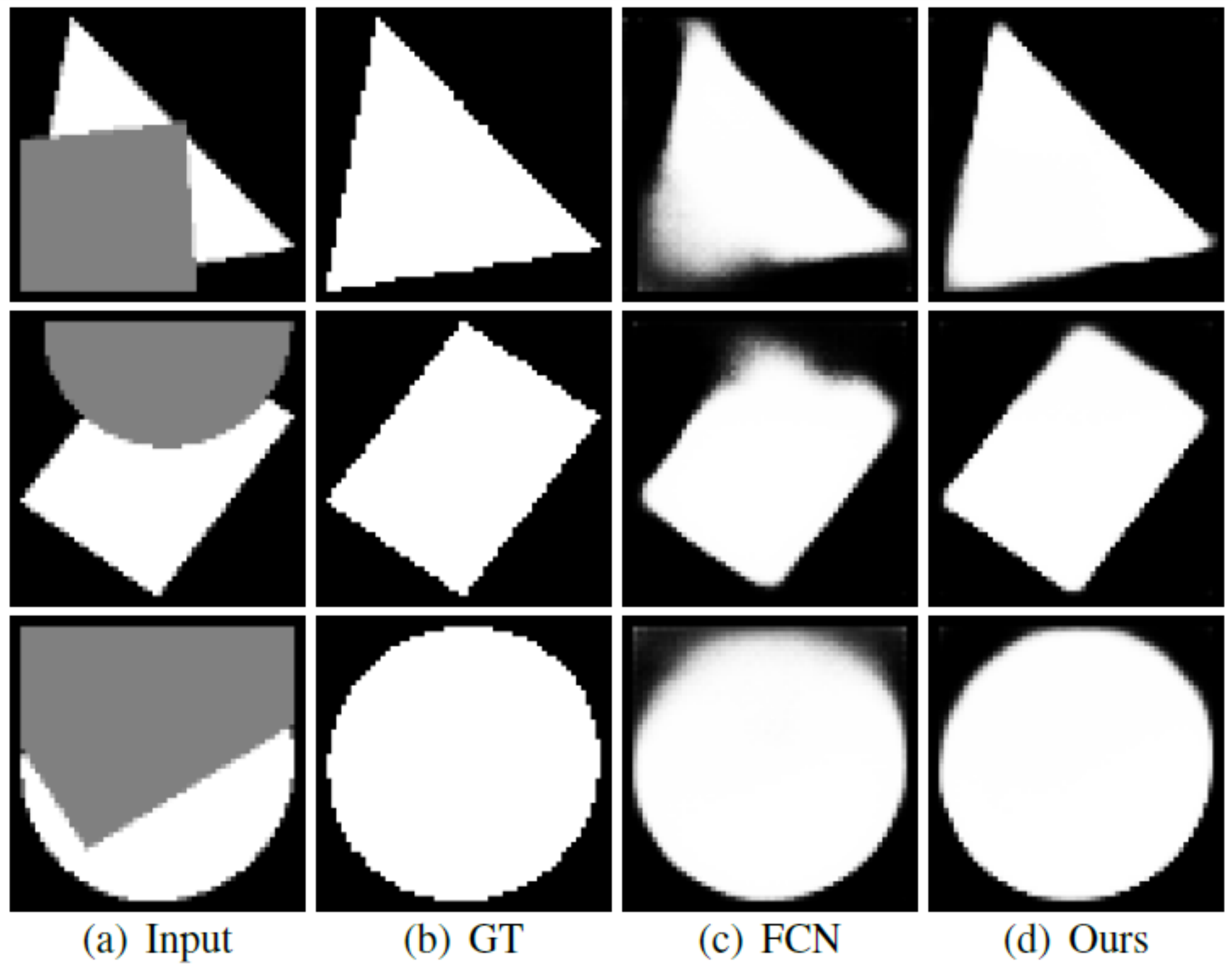}
    \caption{Comparison of shape completion abilities of our algorithm and FCN.}
    \label{fig:occlusion}
\end{figure}

\begin{figure*}[t]
    \centering
    \includegraphics[width=\textwidth]{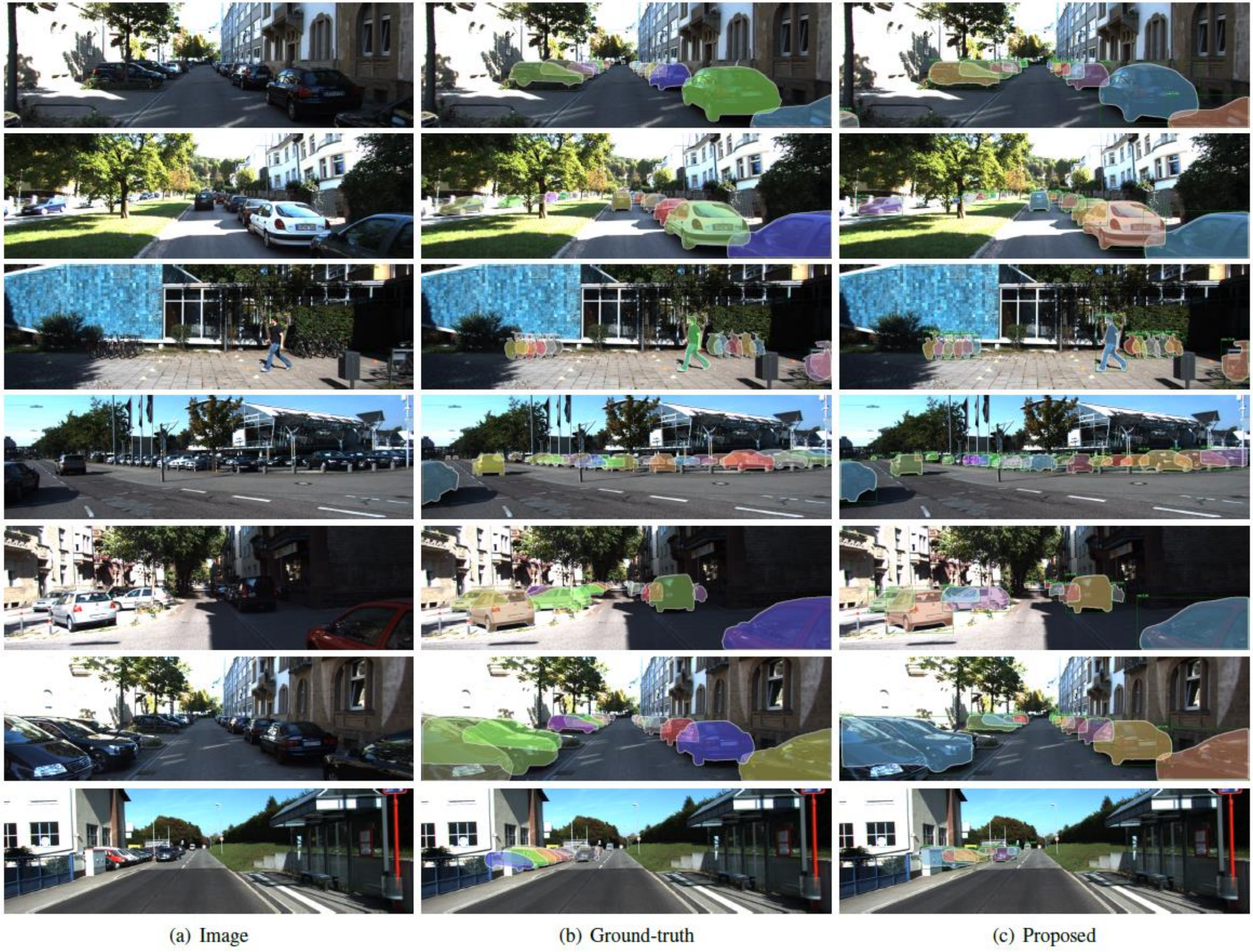}
    \caption{\textbf{Results on KINS dataset~\cite{alhaija2018augmented}}. The segments are depicted by coloured regions. The object masks are generated using the Mask R-CNN pipeline. The last two rows display failure cases of the proposed method.}
    \label{fig:kins_result}
\end{figure*}

To examine the learned vector quantized shape code and the model design choices, we conduct controlled experiments on a synthetic shape dataset. 

\bfsection{Data} We synthesize a database of images containing triangles, rectangles, and ellipses. Each image is 224$\times$224, has up to 9 objects in random positions and orientations, with each object set to a random color, and with a random background color. We generate 5,000 training images and 1,000 evaluation images.





\bfsection{Comparing with VAE}
We use the Mask-RCNN pipeline and compare different shape modeling from FCN (no latent code), VAE (continuous latent code) and VQ-VAE (discrete latent code) modules.

Table~\ref{table:synthesized} compares mAP indices of the three methods on the synthesized dataset. The proposed VQ-VAE method is better able to delineate complete shapes versus the other two methods. Especially, there is a considerable margin between the proposed algorithm and FCN. 

Fig.~\ref{fig:synthesized} shows segmentation results for partially-visible object segmentation. The proposed algorithm discovers these objects; though sometimes the baseline methods fail to predict full geometric shapes. We provide more segmentation results in the supplementary materials.

\bfsection{Robustness to Occlusion}
We assess the occlusion handling abilities of FCN and VQ-VAE when the perfect bounding boxes are given. To this end, we design the following \emph{in silico} psychophysics experiment. 
Given an input shape, we add another shape in front with different degrees of occlusion. We feed these test images into the trained amodal segmentation models with FCN or VQ-VAE modules. 

We quantitatively compare the shape completion performance of the proposed algorithm and FCN. Our algorithm (0.968) outperforms FCN (0.935) in terms of IoU. Moreover, as shown in Fig.~\ref{fig:occlusion}, the proposed algorithm reliably complete the full shapes of occluded objects. However, FCN makes soft predictions on unseen regions, thus restored regions are blurry.







\subsection{Additional Results on Natural Images}



\bgroup
\def\arraystretch{1}
\begin{table}[t]
    \centering
    \caption{Comparison of mAP metric on the KINS dataset~\cite{qi2019amodal}. 
    }
    \vspace{0.5em}
    \begin{tabular}{l|cc}
    \hline 
    ~Detection& ~FCN~&~VQ-VAE (ours)~\\
    \hline
    ~Mask R-CNN~ &  ~0.293~\cite{he2017mask} & \bf{0.303} \\
    ~Mask R-CNN + ASN~ & ~0.311~\cite{qi2019amodal} & \bf{0.315} \\
    \hline
    \end{tabular}
\label{table:kins}
\end{table}
\egroup


To demonstrate the general applicability of our proposed method, we test on an amodal instance segmentation dataset for natural images with a greater diversity of shapes.

\bfsection{Data}
The KINS dataset~\cite{qi2019amodal} is a benchmark for amodal instance segmentation algorithms, which is originally from the KITTI dataset~\cite{geiger2012we}. It consists of 7,474 training and 7,517 test images of driving scenes. The annotated objects belong to one of 7 object classes: pedestrian, cyclist, car, van, tram, truck, and misc-vehicle. The KINS dataset provides both amodal and inmodal ground-truth annotations.

\bfsection{Results}
Table~\ref{table:kins} lists mean average precision metrics of the results of the proposed algorithm with Mask R-CNN~\cite{he2017mask} and Mask R-CNN + ASN~\cite{qi2019amodal}. Our proposed algorithm performs better than the conventional FCN method on the Mask R-CNN pipeline and yields a slightly better result on Mask R-CNN + ASN. Qualitatively, it finds complete masks of occluded cars (Fig.~\ref{fig:kins_result}). In the last two rows in Fig.~\ref{fig:kins_result}, the proposed method fails to segment out cars on the left. This is because the non-maximum suppression removes highly overlapped bounding boxes.

\bgroup
\def\arraystretch{1}
\begin{table}[t]
    \vspace{0.5cm}
    \centering
    \caption{Ablation study on the KINS dataset~\cite{qi2019amodal}.
    }
    \vspace{0.1cm}
    \begin{tabular}{l|c}
    \hline 
    ~Setting~ & ~mAP~\\
    \hline
    ~VQ-VAE~ &~0.281~\\
    ~VQ-VAE + Refinement~ &~0.298~\\
    ~VQ-VAE + Refinement + Occlusion map~ &~\bf{0.303}~\\
    \hline
    \end{tabular}
\label{table:ablation}
\end{table}
\egroup

\bfsection{Ablation Study}
We perform two ablation studies on the KINS dataset. We chose KINS over the embryo dataset for more general analysis, since the objects in KINS have more diverse shapes. We use Mask R-CNN in these studies. First, we remove the occlusion detection branch (VQ-VAE + Refinement). To this end, we train the network without the loss function for occlusion detection $\mathcal{L}_\textrm{o}$. Second, we exclude the refinement decoder in the segmentation module (VQ-VAE). To train the network without the refinement decoder, we minimized the embedding loss $\mathcal{L}_\textrm{e}$ only. 
We compare these two settings with the full architecture (VQ-VAE + Refinement + Occlusion map) on the KINS dataset. Table~\ref{table:ablation} lists the mAP scores for each ablation setting. Our full architecture performs 0.303 mAP, which is better than the other settings. It indicates that all our components are necessary for accurate amodal segmentation. The inferior performance of the setting without refinement comes from the lack of low-level features.

\section{Conclusion}

We proposed an image segmentation method for blastomere instances, which outputs complete masks of cells automatically. The proposed algorithm predicts bounding boxes first and then generates masks. We show that it is effective to learn a mapping from the bounding box features to a shape prior embedding space from a VQ-VAE. This allows us to cope with translucent cells. We also show the benefits of occlusion detection for amodal object detection and segmentation. 
Our method is applicable for any partially visible objects, not only cells but also geometric shapes, cars, or pedestrians.
Experimental results on the embryo, synthesized, and KINS demonstrated that our proposed algorithm outperforms state-of-the-art object instance segmentation methods~\cite{he2017mask,qi2019amodal}.

Our future works include application to other objects in natural scenes and expanding to biomedical problems that suffer occlusions, such as human blood cell segmentation. We also suggest proposal-free amodal segmentation networks with the center prediction to achieve real-time running speed. Lastly, by adopting generative adversarial networks~\cite{goodfellow2014generative}, we might be able to learn shape priors better.

\bibliographystyle{IEEEtran}
\bibliography{tmi_bibliography}

\begin{thebibliography}{10}
\providecommand{\url}[1]{#1}
\csname url@samestyle\endcsname
\providecommand{\newblock}{\relax}
\providecommand{\bibinfo}[2]{#2}
\providecommand{\BIBentrySTDinterwordspacing}{\spaceskip=0pt\relax}
\providecommand{\BIBentryALTinterwordstretchfactor}{4}
\providecommand{\BIBentryALTinterwordspacing}{\spaceskip=\fontdimen2\font plus
\BIBentryALTinterwordstretchfactor\fontdimen3\font minus
  \fontdimen4\font\relax}
\providecommand{\BIBforeignlanguage}[2]{{%
\expandafter\ifx\csname l@#1\endcsname\relax
\typeout{** WARNING: IEEEtran.bst: No hyphenation pattern has been}%
\typeout{** loaded for the language `#1'. Using the pattern for}%
\typeout{** the default language instead.}%
\else
\language=\csname l@#1\endcsname
\fi
#2}}
\providecommand{\BIBdecl}{\relax}
\BIBdecl

\bibitem{armstrong2019Timelapse}
S.~Armstrong, P.~Bhide, V.~Jordan, A.~Pacey, J.~Marjoribanks, and C.~Farquhar,
  ``Time-lapse systems for embryo incubation and assessment in assisted
  reproduction,'' \emph{Cochrane Database of Systematic Reviews}, no.~5, 2019,
  publisher: John Wiley \& Sons, Ltd.

\bibitem{racowsky2011National}
C.~Racowsky, J.~E. Stern, W.~E. Gibbons, B.~Behr, K.~O. Pomeroy, and J.~D.
  Biggers, ``National collection of embryo morphology data into {Society} for
  {Assisted} {Reproductive} {Technology} {Clinic} {Outcomes} {Reporting}
  {System}: associations among day 3 cell number, fragmentation and blastomere
  asymmetry, and live birth rate,'' \emph{Fertility and Sterility}, vol.~95,
  no.~6, pp. 1985--1989, 2011, publisher: Elsevier.

\bibitem{rad2018hybrid}
R.~M. Rad, P.~Saeedi, J.~Au, and J.~Havelock, ``A hybrid approach for multiple
  blastomeres identification in early human embryo images,'' \emph{Computers in
  biology and medicine}, vol. 101, pp. 100--111, 2018, publisher: Elsevier.

\bibitem{sidhu2019Automated}
S.~S. Sidhu and J.~K. Mills, ``Automated {Blastomere} {Segmentation} for
  {Early}-{Stage} {Embryo} {Using} {3D} {Imaging} {Techniques},'' in \emph{2019
  {IEEE} {International} {Conference} on {Mechatronics} and {Automation}
  ({ICMA})}, Aug. 2019, pp. 1588--1593, iSSN: 2152-7431.

\bibitem{moradirad2019CellNet}
R.~Moradi~Rad, P.~Saeedi, J.~Au, and J.~Havelock, ``Cell-{Net}: {Embryonic}
  {Cell} {Counting} and {Centroid} {Localization} via {Residual} {Incremental}
  {Atrous} {Pyramid} and {Progressive} {Upsampling} {Convolution},'' \emph{IEEE
  Access}, vol.~7, pp. 81\,945--81\,955, 2019.

\bibitem{kheradmand2019Preimplantation}
\BIBentryALTinterwordspacing
S.~Kheradmand, P.~Saeedi, J.~Au, and J.~Havelock, ``Preimplantation
  {Blastomere} {Boundary} {Identification} in {HMC} {Microscopic} {Images} of
  {Early} {Stage} {Human} {Embryos},'' in \emph{{arXiv}:1910.05972 [cs, eess,
  q-bio]}, Oct. 2019, arXiv: 1910.05972. [Online]. Available:
  \url{http://arxiv.org/abs/1910.05972}
\BIBentrySTDinterwordspacing

\bibitem{khan2016Segmentation}
\BIBentryALTinterwordspacing
A.~Khan, S.~Gould, and M.~Salzmann, ``\BIBforeignlanguage{en}{Segmentation of
  developing human embryo in time-lapse microscopy},'' in
  \emph{\BIBforeignlanguage{en}{2016 {IEEE} 13th {International} {Symposium} on
  {Biomedical} {Imaging} ({ISBI})}}.\hskip 1em plus 0.5em minus 0.4em\relax
  Prague, Czech Republic: IEEE, Apr. 2016, pp. 930--934. [Online]. Available:
  \url{http://ieeexplore.ieee.org/document/7493417/}
\BIBentrySTDinterwordspacing

\bibitem{li2017fully}
Y.~Li, H.~Qi, J.~Dai, X.~Ji, and Y.~Wei, ``Fully convolutional instance-aware
  semantic segmentation,'' in \emph{{CVPR}}, 2017, pp. 2359--2367.

\bibitem{he2017mask}
K.~He, G.~Gkioxari, P.~Dollár, and R.~Girshick, ``Mask r-cnn,'' in
  \emph{{ICCV}}, 2017, pp. 2961--2969.

\bibitem{liu2018path}
S.~Liu, L.~Qi, H.~Qin, J.~Shi, and J.~Jia, ``Path aggregation network for
  instance segmentation,'' in \emph{{CVPR}}, 2018, pp. 8759--8768.

\bibitem{dai2016instance}
J.~Dai, K.~He, and J.~Sun, ``Instance-aware semantic segmentation via
  multi-task network cascades,'' in \emph{{CVPR}}, 2016, pp. 3150--3158.

\bibitem{knoll1986recognizing}
T.~Knoll and R.~Jain, ``Recognizing partially visible objects using feature
  indexed hypotheses,'' \emph{IEEE Journal on Robotics and Automation}, vol.~2,
  no.~1, pp. 3--13, 1986, publisher: IEEE.

\bibitem{Cootes1995}
\BIBentryALTinterwordspacing
T.~F. Cootes, C.~J. Taylor, D.~H. Cooper, and J.~Graham, ``Active shape models
  \& their training and application,'' \emph{Comput. Vis. Image Underst.},
  vol.~61, no.~1, pp. 38--59, Jan. 1995, number of pages: 22 Publisher:
  Elsevier Science Inc. tex.acmid: 206547 tex.address: New York, NY, USA
  tex.issue\_date: Jan. 1995. [Online]. Available:
  \url{http://dx.doi.org/10.1006/cviu.1995.1004}
\BIBentrySTDinterwordspacing

\bibitem{parizi2014automatic}
S.~N. Parizi, A.~Vedaldi, A.~Zisserman, and P.~Felzenszwalb, ``Automatic
  discovery and optimization of parts for image classification,'' in
  \emph{{arXiv}}, 2014, arXiv: 1412.6598 [cs.CV].

\bibitem{winn2006layout}
J.~Winn and J.~Shotton, ``The layout consistent random field for recognizing
  and segmenting partially occluded objects,'' in \emph{2006 {IEEE} computer
  society conference on computer vision and pattern recognition ({CVPR}'06)},
  vol.~1, 2006, pp. 37--44, tex.organization: IEEE.

\bibitem{li2016amodal}
K.~Li and J.~Malik, ``Amodal instance segmentation,'' in \emph{{ECCV}}, 2016,
  pp. 677--693, tex.organization: Springer.

\bibitem{zhu2017semantic}
Y.~Zhu, Y.~Tian, D.~Metaxas, and P.~Dollár, ``Semantic amodal segmentation,''
  in \emph{{CVPR}}.\hskip 1em plus 0.5em minus 0.4em\relax IEEE, 2017, pp.
  1464--1472.

\bibitem{ehsani2018segan}
K.~Ehsani, R.~Mottaghi, and A.~Farhadi, ``{SeGAN}: {Segmenting} and generating
  the invisible,'' in \emph{{CVPR}}.\hskip 1em plus 0.5em minus 0.4em\relax
  IEEE, 2018, pp. 6144--6153.

\bibitem{follmann2019learning}
P.~Follmann, R.~K. Nig, P.~H. Rtinger, M.~Klostermann, and T.~B. Ttger,
  ``Learning to see the invisible: {End}-to-end trainable amodal instance
  segmentation,'' in \emph{{IEEE} winter conference on applications of computer
  vision ({WACV})}, 2019, pp. 1328--1336, tex.organization: IEEE.

\bibitem{hu2019sail}
Y.-T. Hu, H.-S. Chen, K.~Hui, J.-B. Huang, and A.~G. Schwing, ``{SAIL}-{VOS}:
  {Semantic} amodal instance level video object segmentation-a synthetic
  dataset and baselines,'' in \emph{{CVPR}}, 2019, pp. 3105--3115.

\bibitem{qi2019amodal}
L.~Qi, L.~Jiang, S.~Liu, X.~Shen, and J.~Jia, ``Amodal instance segmentation
  with {KINS} dataset,'' in \emph{{CVPR}}, 2019, pp. 3014--3023.

\bibitem{kingma2013auto}
D.~P. Kingma and M.~Welling, ``Auto-encoding variational bayes,'' in
  \emph{{ICLR}}, 2013.

\bibitem{alhaija2018augmented}
H.~A. Alhaija, S.~K. Mustikovela, L.~Mescheder, A.~Geiger, and C.~Rother,
  ``Augmented reality meets computer vision: {Efficient} data generation for
  urban driving scenes,'' \emph{International Journal of Computer Vision}, vol.
  126, no.~9, pp. 961--972, 2018, publisher: Springer.

\bibitem{molnar2016accurate}
C.~Molnar, I.~H. Jermyn, Z.~Kato, V.~Rahkama, P.~Östling, P.~Mikkonen,
  V.~Pietiäinen, and P.~Horvath, ``Accurate morphology preserving segmentation
  of overlapping cells based on active contours,'' \emph{Scientific reports},
  vol.~6, p. 32412, 2016, publisher: Nature Publishing Group.

\bibitem{Kass1988}
\BIBentryALTinterwordspacing
M.~Kass, A.~Witkin, and D.~Terzopoulos, ``Snakes: {Active} contour models,''
  \emph{International Journal of Computer Vision}, vol.~1, no.~4, pp. 321--331,
  Jan. 1988. [Online]. Available: \url{https://doi.org/10.1007/BF00133570}
\BIBentrySTDinterwordspacing

\bibitem{plissiti2012overlapping}
M.~E. Plissiti and C.~Nikou, ``Overlapping cell nuclei segmentation using a
  spatially adaptive active physical model,'' \emph{IEEE Transactions on Image
  Processing}, vol.~21, no.~11, pp. 4568--4580, 2012, publisher: IEEE.

\bibitem{LeeKim2016}
H.~Lee and J.~Kim, ``Segmentation of overlapping cervical cells in microscopic
  images with superpixel partitioning and cell-wise contour refinement,'' in
  \emph{2016 {IEEE} conference on computer vision and pattern recognition
  workshops ({CVPRW})}, Jun. 2016, pp. 1367--1373, iSSN: 2160-7516.

\bibitem{Bohm2018}
A.~Böhm, A.~Ücker, T.~Jäger, O.~Ronneberger, and T.~Falk, ``{ISOODL}:
  {Instance} segmentation of overlapping biological objects using deep
  learning,'' in \emph{2018 {IEEE} 15th international symposium on biomedical
  imaging ({ISBI} 2018)}, Apr. 2018, pp. 1225--1229, iSSN: 1945-8452.

\bibitem{kihara2016shadows}
Y.~Kihara, M.~Soloviev, and T.~Chen, ``In the shadows, shape priors shine:
  {Using} occlusion to improve multi-region segmentation,'' in \emph{{CVPR}},
  2016, pp. 392--401.

\bibitem{eslami2014shape}
S.~A. Eslami, N.~Heess, C.~K. Williams, and J.~Winn, ``The shape boltzmann
  machine: {A} strong model of object shape,'' \emph{International Journal of
  Computer Vision}, vol. 107, no.~2, pp. 155--176, 2014, publisher: Springer.

\bibitem{ren2015faster}
S.~Ren, K.~He, R.~Girshick, and J.~Sun, ``Faster {R}-{CNN}: {Towards} real-time
  object detection with region proposal networks,'' in \emph{{NIPS}}, 2015, pp.
  91--99.

\bibitem{isack2018k}
H.~Isack, L.~Gorelick, K.~Ng, O.~Veksler, and Y.~Boykov, ``K-convexity shape
  priors for segmentation,'' in \emph{{ECCV}}, 2018, pp. 36--51.

\bibitem{veksler2008star}
O.~Veksler, ``Star shape prior for graph-cut image segmentation,'' in
  \emph{{ECCV}}, 2008, pp. 454--467, tex.organization: Springer.

\bibitem{gulshan2010geodesic}
V.~Gulshan, C.~Rother, A.~Criminisi, A.~Blake, and A.~Zisserman, ``Geodesic
  star convexity for interactive image segmentation,'' in \emph{{CVPR}}, 2010,
  pp. 3129--3136, tex.organization: IEEE.

\bibitem{isack2016hedgehog}
H.~Isack, O.~Veksler, M.~Sonka, and Y.~Boykov, ``Hedgehog shape priors for
  multi-object segmentation,'' in \emph{{CVPR}}, 2016, pp. 2434--2442.

\bibitem{gorelick2017convexity}
L.~Gorelick, O.~Veksler, Y.~Boykov, and C.~Nieuwenhuis, ``Convexity shape prior
  for binary segmentation,'' \emph{IEEE Transactions on Pattern Analysis and
  Machine Intelligence}, vol.~39, no.~2, pp. 258--271, 2017, publisher: IEEE.

\bibitem{wu2016single}
J.~Wu, T.~Xue, J.~J. Lim, Y.~Tian, J.~B. Tenenbaum, A.~Torralba, and W.~T.
  Freeman, ``Single image {3D} interpreter network,'' in \emph{{ECCV}}, 2016,
  pp. 365--382, tex.organization: Springer.

\bibitem{wu2018learning}
J.~Wu, C.~Zhang, X.~Zhang, Z.~Zhang, W.~T. Freeman, and J.~B. Tenenbaum,
  ``Learning shape priors for single-view 3d completion and reconstruction,''
  in \emph{{ECCV}}, 2018, pp. 646--662.

\bibitem{dai2017shape}
A.~Dai, C.~Ruizhongtai~Qi, and M.~Nießner, ``Shape completion using
  {3D}-encoder-predictor {CNNs} and shape synthesis,'' in \emph{{CVPR}}, 2017,
  pp. 5868--5877.

\bibitem{stutz2018learning}
D.~Stutz and A.~Geiger, ``Learning {3D} shape completion from laser scan data
  with weak supervision,'' in \emph{{CVPR}}, 2018, pp. 1955--1964.

\bibitem{kuo_shapemask_2019}
W.~Kuo, A.~Angelova, J.~Malik, and T.-Y. Lin, ``Shapemask: {Learning} to
  segment novel objects by refining shape priors,'' in \emph{Proceedings of the
  {IEEE} {International} {Conference} on {Computer} {Vision}}, 2019, pp.
  9207--9216.

\bibitem{agustsson2017soft}
E.~Agustsson, F.~Mentzer, M.~Tschannen, L.~Cavigelli, R.~Timofte, L.~Benini,
  and L.~V. Gool, ``Soft-to-hard vector quantization for end-to-end learning
  compressible representations,'' in \emph{Advances in neural information
  processing systems}, 2017, pp. 1141--1151.

\bibitem{theis2017lossy}
L.~Theis, W.~Shi, A.~Cunningham, and F.~Huszár, ``Lossy image compression with
  compressive autoencoders,'' \emph{arXiv preprint arXiv:1703.00395}, 2017.

\bibitem{van2017neural}
A.~van~den Oord, O.~Vinyals, and {others}, ``Neural discrete representation
  learning,'' in \emph{Advances in neural information processing systems},
  2017, pp. 6306--6315.

\bibitem{razavi2019generating}
A.~Razavi, A.~v.~d. Oord, and O.~Vinyals, ``Generating diverse high-fidelity
  images with {VQ}-{VAE}-2,'' \emph{arXiv preprint arXiv:1906.00446}, 2019.

\bibitem{he2016deep}
K.~He, X.~Zhang, S.~Ren, and J.~Sun, ``Deep residual learning for image
  recognition,'' in \emph{{CVPR}}, 2016, pp. 770--778.

\bibitem{deng2009imagenet}
J.~Deng, W.~Dong, R.~Socher, L.-J. Li, K.~Li, and L.~Fei-Fei, ``{ImageNet}: {A}
  large-scale hierarchical image database,'' in \emph{{CVPR}}, 2009, pp.
  248--255, tex.organization: Ieee.

\bibitem{lin2014microsoft}
T.-Y. Lin, M.~Maire, S.~Belongie, J.~Hays, P.~Perona, D.~Ramanan, P.~Dollár,
  and C.~L. Zitnick, ``Microsoft {COCO}: {Common} objects in context,'' in
  \emph{{ECCV}}, 2014, pp. 740--755, tex.organization: Springer.

\bibitem{khan_deep_2016}
A.~Khan, S.~Gould, and M.~Salzmann, ``Deep convolutional neural networks for
  human embryonic cell counting,'' in \emph{European {Conference} on {Computer}
  {Vision} workshops}.\hskip 1em plus 0.5em minus 0.4em\relax Springer, 2016,
  pp. 339--348.

\bibitem{geiger2012we}
A.~Geiger, P.~Lenz, and R.~Urtasun, ``Are we ready for autonomous driving? the
  {KITTI} vision benchmark suite,'' in \emph{{CVPR}}, 2012, pp. 3354--3361,
  tex.organization: IEEE.

\bibitem{goodfellow2014generative}
I.~Goodfellow, J.~Pouget-Abadie, M.~Mirza, B.~Xu, D.~Warde-Farley, S.~Ozair,
  A.~Courville, and Y.~Bengio, ``Generative adversarial nets,'' in
  \emph{{NIPS}}, 2014, pp. 2672--2680.

\end{thebibliography}

\end{document}